\documentclass[graybox]{svmult}

\usepackage{type1cm}        
\usepackage{makeidx}         
\usepackage{graphicx}        
\usepackage{multicol}        
\usepackage[bottom]{footmisc}

\usepackage{newtxtext}       %
\usepackage{newtxmath}       

\usepackage{amsmath}

\usepackage[justification=centering]{caption}

\usepackage{geometry}
\usepackage[section]{placeins}

\makeindex             

\begin{document}

\title*{Modeling Fuzzy Cluster Transitions for Topic Tracing}

\author{Xiaonan Jing, Yi Zhang, Qingyuan Hu, Julia Taylor Rayz}
\institute{Xiaonan Jing \at Purdue University, West Lafayette, IN, \email{jing@purdue.edu}
\and Yi Zhang \at Purdue University, West Lafayette, IN \email{zhan3050@purdue.edu}
\and Qingyuan Hu \at Purdue University, West Lafayette, IN \email{hu528@purdue.edu}
\and Julia Taylor Rayz \at Purdue University, West Lafayette, IN \email{jtaylor1@purdue.edu}}
%
%
\maketitle

\abstract{Twitter can be viewed as a data source for Natural Language Processing (NLP) tasks. The continuously updating data streams on Twitter make it challenging to trace real-time topic evolution. In this paper, we propose a framework for modeling fuzzy transitions of topic clusters. We extend our previous work on crisp cluster transitions by incorporating fuzzy logic in order to enrich the underlying structures identified by the framework. We apply the methodology to both computer generated clusters of nouns from tweets and human tweet annotations. The obtained fuzzy transitions are compared with the crisp transitions, on both computer generated clusters and human labeled topic sets.}

\section{Introduction}
\label{sec:intro}
Real-time event identification can be challenging due to the constantly updating data streams. Such identification is crucial, however, for time-sensitive events, such as natural disasters or any unrests that may require an understanding of emergent sub-events. For a real-time event $T$ and a set of timepoints $\{t_1, t_2, ..., t_n\}$, a temporal (sub-)event can be characterized by a state $s_i$ at timepoint $t_i (i < n)$, where the state $s_i$ consists of a set of features of  $T$. As $t_i$ moves along the time axis, the state $s_i$ may also encompass change---an anomaly or an additional event that makes the state differ from the previous one. Such change is often referred to as a "transition" between $s_i$ and $s_{i+1}$. Previously, stream-based detection has focused on detecting a new state which induces a change \cite{guille2015event,meladianos2015degeneracy,fedoryszak2019real}. However, the transition itself which describes how a state can change is equally important to trace the semantic topic evolution. We are interested in detecting the topic transitions in a continuous time space to learn how a dynamic event unfolds. In this paper, we investigate fuzzy \cite{zadeh1965} transitions between events: while it may seem like a good idea to treat any transition as binary at first, we believe that fuzzy transitions show information about sub-events. For instance, a change of topic from one event to another is usually gradual, and modeling it as a simple "different or not" is not the most adequate solution. 

In our previous work, we proposed to employ a graph structure to model temporal states in identifying cluster transitions \cite{jing2021tracing}. A temporal Graph-of-Words (GoW), $G=(V, E)$ was constructed for textual data at each timepoint, with vertices $V$ being unique tokens (nouns and named entities) and edges $E$ being normalized point-wise mutual information between the tokens. Markov Clustering algorithm \cite{van2008graph} was applied on each temporal GoW to obtain topic clusters. Finally, we proposed a cluster transition model to trace the topic evolution and duration on a global timescale. Our framework was capable of tracing the topic progression as compared to the progression generated through human labeled topic sets. 

While our framework adopted a crisp cluster transition model which used a $match$ function to determine a unique transition type, the transition assignment, in its nature, should be fuzzy. More precisely, the following two scenarios should be considered when assigning transition types: 1) a single transition type can have strength and is not always the same for different pairs of clusters; 2) the transition is vague as it can be treated as two or more types of transitions. Scenario 1 can be explained by a simple example. For instance, one can argue that a cluster $Cl_a$ at $t_1$ stays \textit{strongly} "unchanged" at $t_2$ if absolutely nothing has changed about $Cl_a$. Similarly, $Cl_a$ can be said to be \textit{relatively} "unchanged" if most of its elements stay through $t_2$. On the other hand, scenario 2 is more of a question that if a single type is sufficient to describe a transition when the conditions for multiple transitions are satisfied. In this paper, we extend our previous framework \cite{jing2021tracing} to introduce fuzzy logic into cluster transition modeling to model both of these scenarios.

\section{Related Work}
\label{sec:related_work}


In recent decades, with the prevalence of social media among people of all ages, short messages such as tweets have attracted many researchers to explore the possibility of identifying events from social media. Clustering is one of the common means to analyze tweet streams, and attempts have been made to use fuzzy-based approaches for tweet event identification. Compared with traditional hard clustering such as K-means, fuzzy-based clustering is more complex, and it provides relatively higher clustering accuracy and more realistic attribution probabilities \cite{alnajran_cluster_2017}. Zadeh et al. \cite{zadeh2015fuzzy} utilized fuzzy clustering to gain insights related to temporal trends and hashtags popularity from Twitter hashtags. As opposed to set crisp boundaries for the categorizations of hashtags, a hashtag can belong to multiple clusters according to certain degrees of membership. However, Zaheh et al. did not apply fuzzy logic to cluster transitions, which is what we intend to highlight in this paper.

In real-time event clustering, clusters can be influenced by the content of the underlying data stream. In order to derive valuable insights from clustering results for a specific time interval, modeling cluster transitions is a critical step in tracking and understanding clustering changes. Spiliopoulou et al. \cite{spiliopoulou_monic_2006} proposed a framework, MONIC, for the modeling and tracking of cluster transitions. A cluster transition at a given timepoint was defined as "a change experienced by a cluster that has been discovered at an earlier timepoint." The authors subsequently named and defined five external transitions and four internal transitions of the cluster. Later, Oliveira and Gama \cite{oliveira_framework_2012} proposed a similar cluster transition model, MEC, which focused more on the external evolution of the cluster structure. They utilized a bipartite graph to visualize the transitions of clusters during a given time interval. Gong et al. \cite{gong_clustering_2017} proposed a density-based stream clustering algorithm, EDMStream, with the assumption that "cluster centers are surrounded by neighbors with lower local density". By monitoring these density mountains, their model could effectively detect the evolution of clusters.

All three mentioned cluster transition models use a matching function with a threshold to identify transitions across consecutive time points for each type. To find the appropriate thresholds for each transition type, Oliveira and Gama\cite{oliveira_framework_2012} analyzed the sensitivity of the thresholds for each transition type, and Gong et al.\cite{gong_clustering_2017} implemented an algorithm to achieve the adaptive adjustment of these thresholds, but none of them chose to fuzzify the cluster transitions.

\section{Modeling Cluster Transition with Fuzzy Logic}
\label{sec:fuzzy_cluster_transition}
%
In this section, we elaborate on how fuzzy logic can be applied to crisp cluster transition types. Due to space limitations, we omit the implementation details of our previous framework \cite{jing2021tracing} and focus on the differences between crisp and fuzzy transition models. 

    \begin{table*}[t]
        \caption{Statistics of the dataset split by day}
        \centering
        \small 
        \begin{tabular}{lccccccccccccccc}
            \hline
            Timepoint & \multicolumn{1}{l}{8/19} & \multicolumn{1}{l}{8/20} & \multicolumn{1}{l}{8/21} & \multicolumn{1}{l}{8/22} & \multicolumn{1}{l}{8/23} & \multicolumn{1}{l}{8/24} & \multicolumn{1}{l}{8/25} & \multicolumn{1}{l}{8/26} & \multicolumn{1}{l}{8/27} & \multicolumn{1}{l}{8/28} & \multicolumn{1}{l}{8/29} & \multicolumn{1}{l}{8/30} & \multicolumn{1}{l}{8/31} & \multicolumn{1}{l}{9/1} & \multicolumn{1}{l}{9/2} \\ \hline
            \# of Tweets & 38 & 89 & 87 & 65 & 27 & 68 & 53 & 29 & 19 & 53 & 23 & 18 & 40 & 35 & 16 \\
            \# of Nodes & 139 & 218 & 167 & 177 & 69 & 189 & 204 & 101 & 65 & 137 & 72 & 56 & 122 & 128 & 113 \\ 
            \# of Clusters & 12 & 27 & 20 & 24 & 9 & 25 & 23 & 14 & 6 & 24 & 8 & 8 & 10 & 14 & 9 \\ 
            \# of labeled sets & 8 & 14 & 10 & 14 & 6 & 6 & 10 & 12 & 9 & 10 & 8 & 7 & 13 & 6 & 12 \\\hline
            \end{tabular}
        \label{tab:dataset}
    \end{table*}
    
The data we model was collected from  Aug 19th to Sep 2nd, representing COVID-19-related tweets from a local community. The dataset was split by Twitter timestamps into 15 timepoints to learn topic transitions across two weeks (Table \ref{tab:dataset}). We first constructed a temporal graph for each timepoint, with nodes being the unique nouns or named entities contained in that timepoint and edge weights being the normalized point-wise mutual information value between the nodes. Each temporal graph was clustered with a modified Markov Clustering algorithm. We validated our approach by comparing the topic transition flows between the computer generated clusters and human-annotated topic sets. Table \ref{tab:dataset} shows the statistics of the dataset and the clustering results on all temporal graphs. The human-annotation was done on each tweet with up to three summarized terms, while the computer generated clusters grouped topics based on nouns and named entities. Thus, a mismatch in the number of clusters exist due to the additional layer of summarization. For elucidation of the implementation and clustering results, please refer to \cite{jing2021tracing}.

\subsection{Crisp Cluster Transitions for Topic Tracing}
\label{sec:sub_cluster_trainsition}
In this section, we briefly describe how the crisp topic transition framework was implemented in our previous work \cite{jing2021tracing} in order to compare it to fuzzy transition framework reported in this paper. After obtaining the temporal clustering for each timepoint, we observed that the common topical elements in clustering at different timepoints could help to determine how clusters progress over time. An intuition is that a cluster $Cl_a$ at timepoint $t_1$ can exhibit several various behaviors, judging by how the common elements change, going into timepoint $t_2$. We modeled these behaviors as transitions between two consecutive timepoints and traced the progression of clusters across all timepoints.

We defined the following crisp pairwise transition types for a cluster $Cl_a$ at timepoint $t_i$ to timepoint $t_{i+1}$: "unchanged", "absorb", "dissolve", "split", "merge", "disappear" and "emerge" \cite{jing2021tracing, spiliopoulou_monic_2006, oliveira_framework_2012}. As shown in Table \ref{tab:cluster_transition_types}, $Cl_a$ represents a cluster at timepoint $t_i$, while $Cl_b$ denotes a corresponding cluster at timepoint $t_{i+1}$. We employed a $match_\alpha$ function with a threshold $\alpha$ to measure the overlap of tokens of two clusters. For a better observation of the cluster transitions through time, pairwise transitions are visualized using a flow (Figures \ref{fig:crisp_clt_computer} and \ref{fig:crisp_clt_human}). In the figures, each cluster is presented as a node and transitions are denoted as edges between the cluster nodes. Different transition types are represented by various colors. Whenever a new cluster emerged, a new transition sequence was created, and the transition types were determined based on the definitions in Table \ref{tab:cluster_transition_types}. 

It should be noted that the threshold $\alpha = 2/3$ was chosen opportunistically for the $match_\alpha$ function to determine if two clusters are the same. Choosing the $\alpha$ value in the crisp version of the transition tracing can be challenging as it serves as the fundamental component to determine the transition quantities, qualities and types. However, the similarity between a pair of clusters is not binary by nature. In the crisp version of the $match_\alpha$ function, the difference between "similar" and "dissimilar" can sometimes be determined by changing one cluster element. The difference, nevertheless, from changing a single element should not be this large. With this intuition, we modify the crisp transitions by adding a soft layer of fuzziness to the $match_\alpha$ function to enrich the transition framework.

    \begin{table}[t]
        \centering
        \caption{Cluster transition types for cluster $Cl_a$ at timepoint $t_i$}
        \begin{tabular}{|c|c|} \hline \\[-1em]
            \textbf{Transition Type} & \textbf{Mathematical Definition} \\ \hline \\[-1em]
            unchanged & $Cl_a \rightarrow Cl_a'$, where $Cl_a' = match_{\alpha}(Cl_a)$ \\ \hline \\[-1em]
            absorb & $Cl_a \rightarrow Cl_b$, where $match_{\alpha}(Cl_a) \subset Cl_b$ and $Cl_b-match_{\alpha}(Cl_a) \not\subset Cl_b $\\ \hline\\[-1em]
            dissolve & $Cl_a \rightarrow Cl_b$, where $match_{\alpha}(Cl_b) \subset Cl_a$ and $Cl_b-match_{\alpha}(Cl_b) \not\subset Cl_a$\\ \hline \\[-1em]
            split & $Cl_a \rightarrow \{{Cl_{b1}, Cl_{b2}, ..., Cl_{bm}}\}$, where $\bigcup_{1}^{m} Cl_{bj} = match_{\alpha}(Cl_a)$ \\ \hline \\[-1em]
            merge & $\{{Cl_{a1}, Cl_{a2}, ..., Cl_{an}}\} \rightarrow Cl_b$, where $\bigcup_1^n Cl_{ai} = match_{\alpha}(Cl_b)$\\ \hline\\[-1em]
            disappear & $Cl_a \rightarrow$ \O \\ \hline\\[-1em]
            emerge & \O\ $ \rightarrow Cl_b$ \\ \hline 
        \end{tabular}
        \label{tab:cluster_transition_types}
    \end{table}

\subsection{Fuzzified Cluster Transitions}
\label{sec:sub_fuzzy_cluster_transition} 
We fuzzify the $match_\alpha$ function defined in Table \ref{tab:cluster_transition_types} by replacing the crisp threshold $\alpha$ with trapezoidal membership functions $\mu$. For transition types "unchanged", "absorbed", "dissolved", "splits", and "merged", the following trapezoidal functions (Equations \eqref{eq:mu_functions}) determine the strength of the transition from cluster $Cl_a$ at timepoint $t_i$ to cluster $Cl_b$ at timepoint $t_{i+1}$. For transition type "disappear", Equations (\ref{eq:mu_functions}) determine the strength of disappearance for cluster $Cl_a$ going from $t_i$ to $t_{i+1}$. It should be noted that the strength of transition type "emerge" is always strong at any timepoint when a new cluster appear. Thus, we do not fuzzify the "emerge" transitions. In Equations (\ref{eq:mu_functions}), $[b, c]$ denotes the core interval, and $[a, d]$ nodes the support limit, with $ 0 < a < b < c < d < 1 $. 

    \begin{subequations}
        \label{eq:mu_functions}
        \begin{align}
        \centering 
            \begin{split}
             \mu_{weak}(x) &= \begin{cases} 
                                1, \quad  x \leq a \\
                                \frac{b-x}{b-a}, \quad a \leq x \leq b \\
                                0, \quad  x \geq b
                                \end{cases} 
            \end{split}\\
            \begin{split}
             \mu_{medium}(x) &= \begin{cases} 
                                0, \quad  x \leq a \enspace or \enspace x \geq d \\ 
                                \frac{x-a}{b-a}, \quad a \leq x \leq b \\ 
                                1, \quad b \leq x \leq c \\ 
                                \frac{d-x}{d-c}, \quad c \leq x \leq d
                                \end{cases}  
            \end{split}\\
            \begin{split}
             \mu_{strong}(x) &= \begin{cases} 
                                0, \quad x \leq c \\ 
                                \frac{x-c}{d-c}, \quad c \leq x \leq d \\ 
                                1, \quad x \geq d 
                                \end{cases} 
            \end{split}
        \end{align}    
    \end{subequations}


For each cluster at timepoint $t_i$, if a cluster $Cl_a$ shares common elements with a cluster $Cl_b$ at $t_{i+1}$, we say a transition exists between $Cl_a$ and $Cl_b$. The discourse $x$, the core (of fuzzy set) interval $[b, c]$, and the support limit [$a$, $d$] are defined as the following: 
    
    \begin{subequations}
        \label{eq:discourse_definitions}
        \begin{align}
        \centering 
            \begin{split}\label{eq:sub1}
            "unchanged"&:  \begin{cases} 
                                x = \frac{|Cl_a \cap Cl_b|}{|Cl_a \cup Cl_b|}, \enspace x \in [0, 1] \\
                                a = 0.3, \enspace b = 0.4, \enspace c = 0.6, \enspace d = 0.7 
                                \end{cases} 
            \end{split}\\
            \begin{split}\label{eq:sub2}
            "absorb"&:  \begin{cases} 
                                x = max(\frac{|Cl_a \cap Cl_b|}{|Cl_a|} - \frac{|Cl_a \cap Cl_b|}{|Cl_a \cup Cl_b|}, \enspace 0), \enspace x \in [0, 1] \\
                                a = 0.3, \enspace b = 0.4, \enspace c = 0.6, \enspace d = 0.7 
                                \end{cases} 
            \end{split}\\
            \begin{split}\label{eq:sub3}
            "dissolve"&:  \begin{cases} 
                                x = max(\frac{|Cl_a \cap Cl_b|}{|Cl_b|} - \frac{|Cl_a \cap Cl_b|}{|Cl_a \cup Cl_b|}, \enspace 0), \enspace x \in [0, 1] \\
                                a = 0.3, \enspace b = 0.4, \enspace c = 0.6, \enspace d = 0.7 
                                \end{cases} 
            \end{split}\\
            \begin{split}\label{eq:sub4}
            "split"&:  \begin{cases} 
                                x = \frac{|\bigcup_1^m(Cl_a \cap Cl_{bj})|}{|\bigcup_1^m(Cl_a \cup Cl_{bj})|}, \enspace x \in [0, 1] \\
                                a = 0.3, \enspace b = 0.4, \enspace c = 0.6, \enspace d = 0.7 
                                \end{cases} 
            \end{split}\\
            \begin{split}\label{eq:sub5}
            "merge"&:  \begin{cases} 
                                x = \frac{|\bigcup_1^n(Cl_{ai} \cap Cl_b)|}{|\bigcup_1^n(Cl_{ai} \cup Cl_b)|}, \enspace x \in [0, 1] \\
                                a = 0.3, \enspace b = 0.4, \enspace c = 0.6, \enspace d = 0.7 
                                \end{cases} 
            \end{split}\\
            \begin{split}\label{eq:sub6}
            "disappear"&:  \begin{cases} 
                                x = \frac{|Cl_a \not\in \, t_{i+1}|}{|Cl_a|}, \enspace x \in [0, 1] \\
                                a = 0.25, \enspace b = 0.35, \enspace c = 0.55, \enspace d = 0.65 
                                \end{cases} 
            \end{split}
        \end{align}    
    \end{subequations}

In Equations \eqref{eq:sub1} - \eqref{eq:sub6}, each cluster is treated as a set to which the general set operations, intersection $x \cap y$, union $x \cup y$, difference $x - y$, and number of elements $|x|$ apply. It should be noted that the types "split" and "merge" involve 1-to-M (1-to-many) mapping of the clusters while the type "unchanged" can be considered as such mapping when M takes the value 1. Thus, we enforce the cluster $Cl_a$ with transition type "split" or "merge" to have element mapping with at least two clusters at $t_{i+1}$ by definition of these types. However, we do not enforce the 1-to-1 mapping for types such as "unchanged" as by definition, a "split" can be treated as multiple \textit{non-strong} "unchanged". Therefore, as long as an intersection exists between cluster $Cl_a$ and $Cl_b$, we consider it for all 1-to-1 types of transition ("unchanged", "absorb", "dissolve"). 

By definition, an "unchanged" cluster $Cl_a$ at $t_{i+1}$ can also be treated as "absorbed" by $Cl_b$. However, a cluster with an "unchanged" membership 1 should not have a high "absorb" membership. To distinguish, we consider the difference between the intersection with respect to $Cl_a$ and the "unchanged". The same can be applied to differentiate the types "unchanged" and "dissolved". As mentioned previously, the "disappear" type exhibits different behavior compared to other types, which involve only one cluster, namely itself, during the transition. Thus, for a cluster $Cl_a$, we define its discourse $x$ as the ratio between the number of elements do not present at timepoint $t_{i+1}$ and the size of $Cl_a$ at $t_i$. Furthermore, we use a different core interval for "disappear," considering common sense intuition for determining the amount of cluster loss needed for it to start to disappear. 

\section{Implementation: Comparing Crisp and Fuzzy Cluster Transitions}
\label{sec:result}

In this section, the transition progression charts generated by the crisp and fuzzy cluster transition models\footnote{The implementation details can be found at https://github.com/lostkuma/TopicTransition} are compared for both computer generated and human annotated clusters. 

\begin{figure}[!htb]
        \centering
        \includegraphics[width=0.95\linewidth]{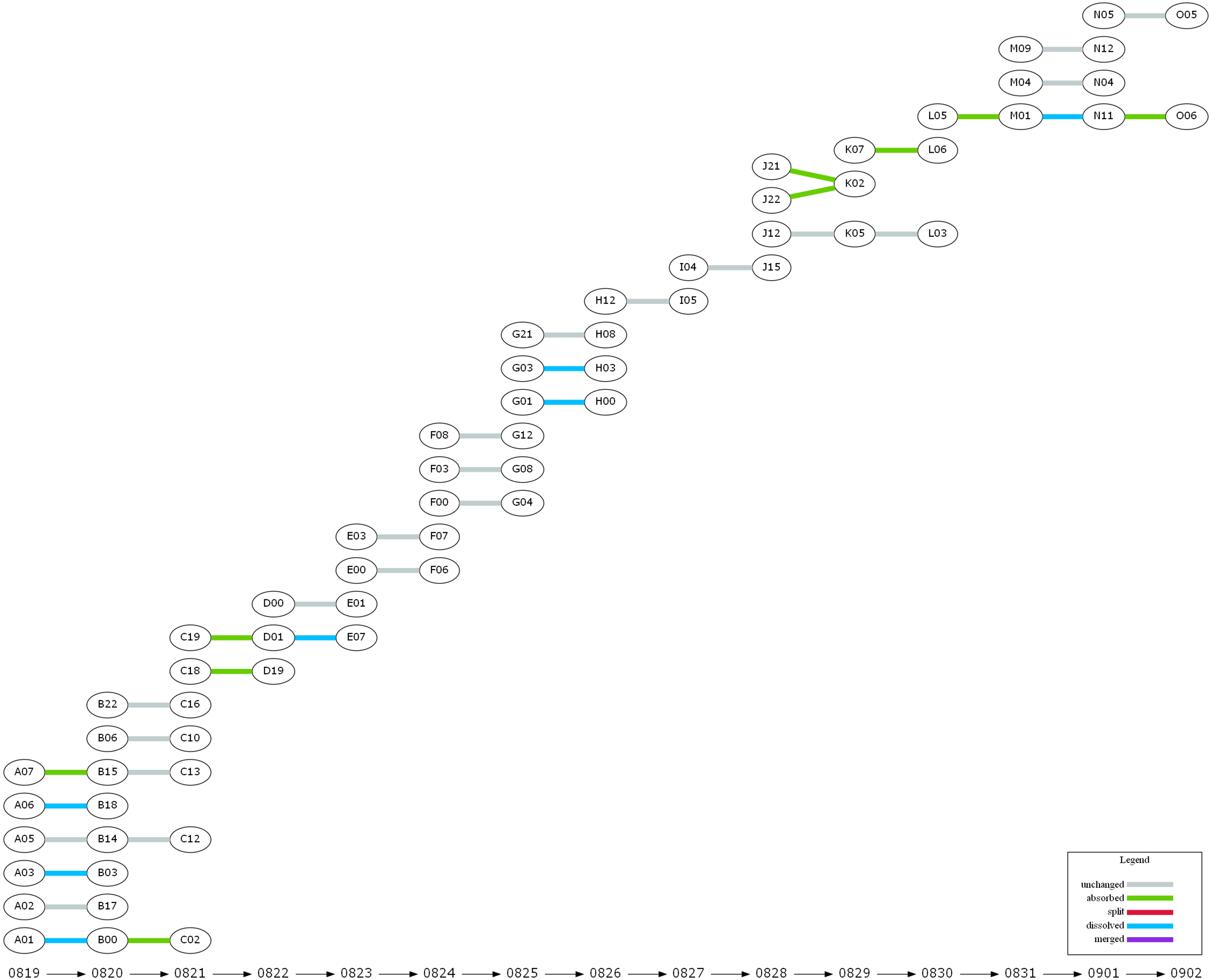}
        \caption{Cluster progression plot for computer generated clusters with crisp transition model}
        \label{fig:crisp_clt_computer}  
    \end{figure}

\begin{figure}[!htb]
        \centering
        \includegraphics[width=0.95\linewidth]{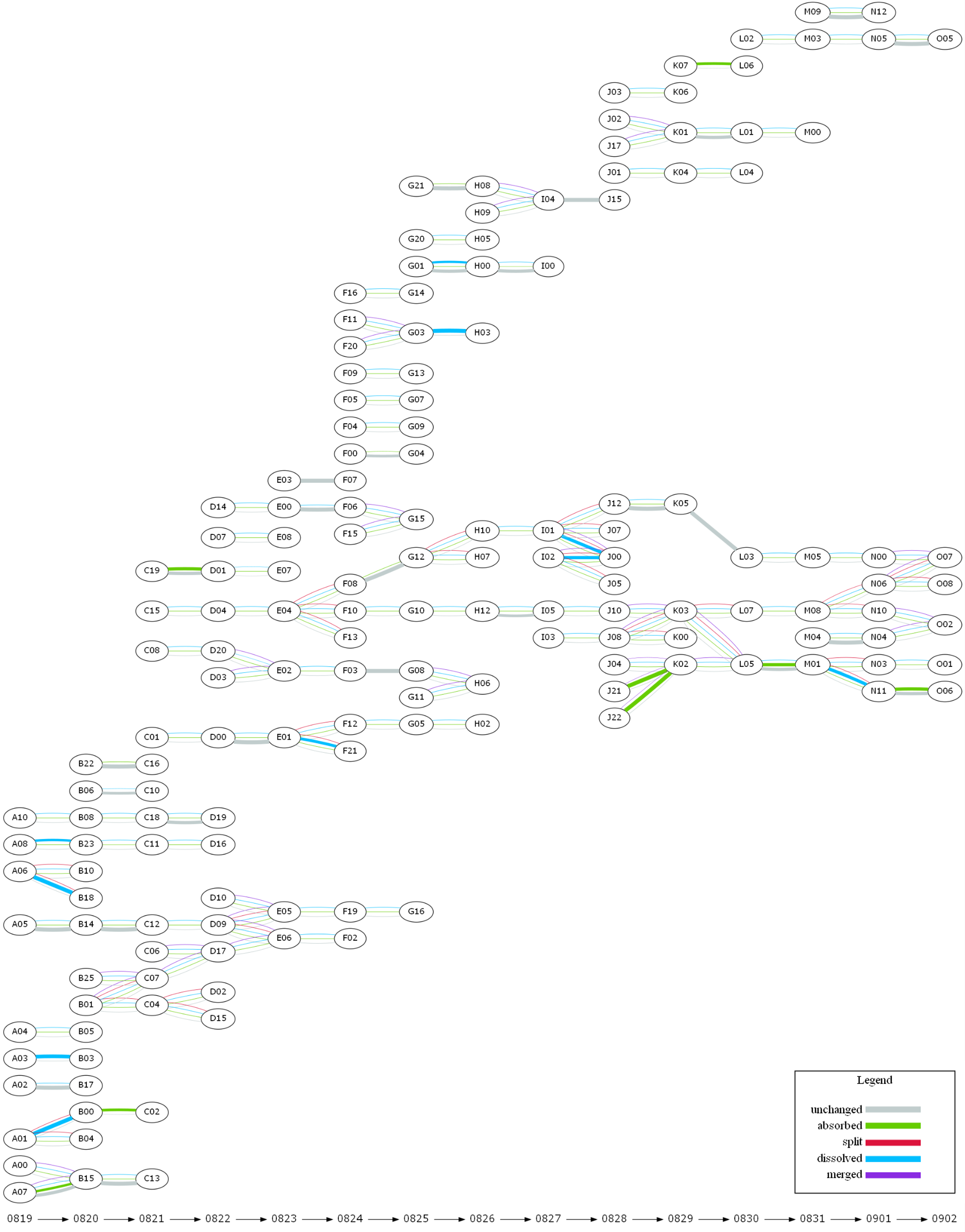}
        \caption{Cluster progression plot for computed clusters with fuzzy transition model}
        \label{fig:fuzzy_clt_computer}      
    \end{figure}

We start with the comparison of the computer generated clusters. Figure \ref{fig:crisp_clt_computer} and Figure \ref{fig:fuzzy_clt_computer} are the flows generated by crisp and fuzzy transition models respectively. Recall that the crisp transition uses a $match_\alpha$ function with a hard threshold to assign a single transition for each pairwise clusters. Furthermore, many pairwise clusters with a smaller number of intersecting elements are excluded by the threshold to be considered as a transition. On the other hand, in Figure \ref{fig:fuzzy_clt_computer}, the fuzzy cluster transitions are shown with edge weight being the membership value $\mu(x)$ of corresponding fuzzy sets (\textit{weak, medium, strong}), with \textit{strong} set shown as a much thicker line than a \textit{weak} one. It should be noted that an edge is assigned multiple fuzzy sets (i.e. \textit{strong} "unchanged", \textit{weak} "absorb") as long as at least one intersecting element exists between the pairwise clusters. 
By comparing the two figures, we can see that 1-to-M transition types ("merge", "split") are not present in the crisp transition chart while they frequently appear in the fuzzy transition chart. Additionally, it can be seen that a large number of the 1-to-M transitions appear with thinner edges (corresponding to the \textit{weak} set) in the fuzzy transitions, which is consistent with the crisp version---the lower membership transition types were excluded by the threshold. Nevertheless, the fuzzy transitions can provide richer information of the underlying structures of the cluster relationships.

\begin{figure} [!htb]
        \centering
        \includegraphics[width=0.95\linewidth]{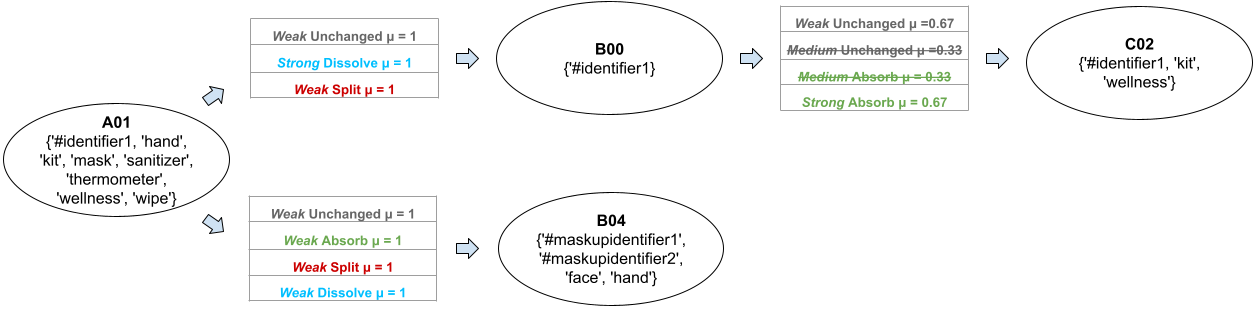}
        \caption{Content of the transition flow starting at cluster $A01$ as shown in Figure \ref{fig:fuzzy_clt_computer}}
        \label{fig:fuzzy_content_example}
    \end{figure}

Figure \ref{fig:fuzzy_content_example} illustrates the contents of an example fuzzy transition flow starting at cluster $A01$ taken from Figure \ref{fig:fuzzy_clt_computer}. The topical elements in each cluster is visualized in the cluster node. The transition types, using the same colored legend, is visualized with their fuzzy set and membership value $\mu$. It should be noted that when multiple strength fuzzy sets present under the same transition type, the fuzzy set with the higher $\mu$ value is selected to present in the transition flow chart. It can be seen in Figure \ref{fig:fuzzy_content_example} that there is one shared element "\#identifier1" going from cluster $B00$ to cluster $C02$, which resulted in two types of transitions -- "unchanged" and "absorb". The transition fuzzy sets were computed using Equations \ref{eq:mu_functions} and \ref{eq:discourse_definitions} to determine the strength of each transition. As a result, a total of 4 fuzzy sets, namely \textit{weak} "unchanged", \textit{medium} "unchanged", \textit{medium} "absorb", and \textit{strong} "absorb" with different $\mu$ values were computed. As the membership $\mu$ takes a larger value at \textit{weak} "unchanged" than \textit{medium} "unchanged" fuzzy set, the prior was taken to characterize this transition. Similar idea applies to the "absorb" fuzzy set. 

\begin{figure}[!htb]
        \centering
        \includegraphics[width=0.95\linewidth]{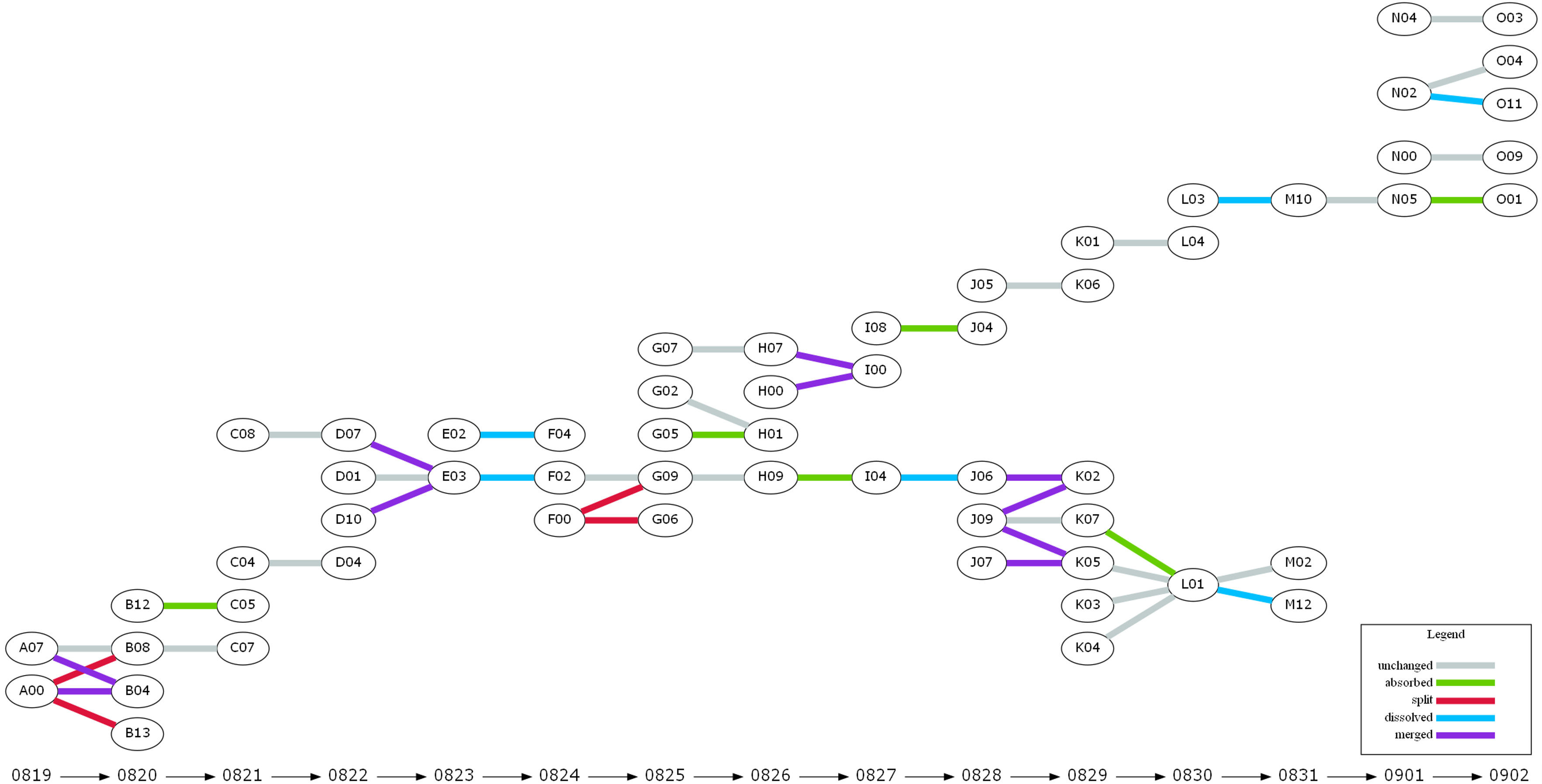}
        \caption{Cluster progression plot for human annotated topics with crisp transition model}
        \label{fig:crisp_clt_human}  
    \end{figure}

\begin{figure}[!htb]
        \centering
        \includegraphics[width=0.95\linewidth]{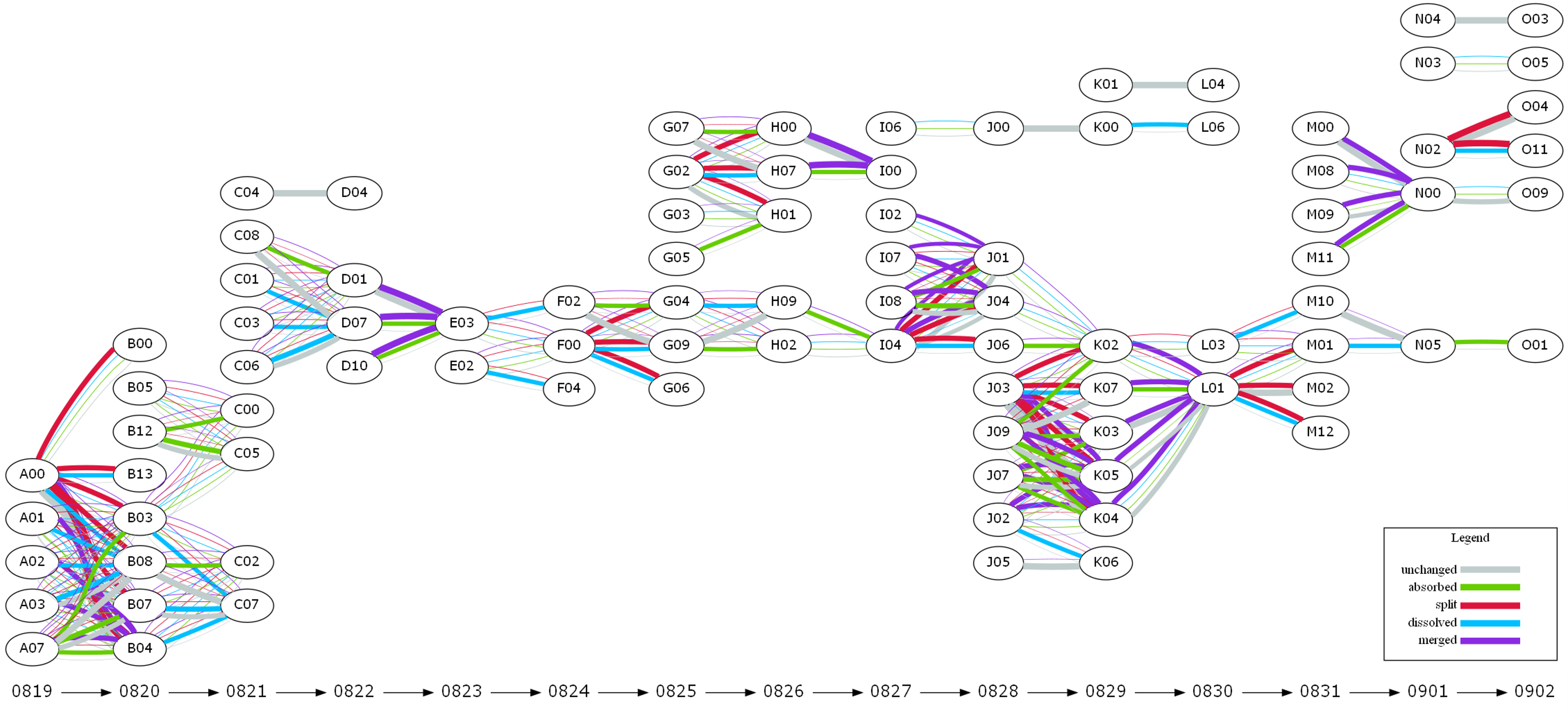}
        \caption{Cluster progression plot for human annotated topics with fuzzy transition model}
        \label{fig:fuzzy_clt_human}     
    \end{figure}

Figure \ref{fig:crisp_clt_human} and Figure \ref{fig:fuzzy_clt_human} illustrate the transition charts generated by crisp and fuzzy transition models on human annotated topic labels. Similar trends with computer generated clusters can be noticed in human annotated clusters regarding the \textit{weak} transitions in Figure \ref{fig:fuzzy_clt_human}. Those transitions tend to disappear when a crisp threshold is applied. Unlike computer generated clusters, the number of nodes present (if the node does not emerge and disappear after the same day) in both human annotated cluster transitions is roughly the same. This is potentially due to the mechanism used for labeling the tweets---the human annotated clusters contain far less labels per cluster and several labels which summarizes the event repeat across various timepoint. The frequently appearing labels add more complexity to the crisp transitions, which caused the fuzzy transition chart to only display more lower membership edges but not a increasingly number of nodes. 

It is more difficult to compare fuzzy human-based transition graph with fuzzy computer-generated transition plot. At the first glance, the graphs are very different. This is not surprising since the annotation used by a human was on a much coarser grain size than that of a computer generated one, yet fuzzy methodology shows even small differences represented by weak transitions. However, one may notice a lot of activity starting from 8/27 in merging and splitting of clusters in both figures. This suggests a gradual change of topic in conversations with call-backs to the original topics, that is picked up by both human annotation and computer-generated clusters. 
    
\section{Conclusion} 
\label{sec:conclusion}
In this paper, we proposed a topic tracing framework by incorporating fuzzy logic into crisp cluster transition modeling. To the best of our knowledge, this is the first work to introduce fuzzy transition types by defining the transition strength fuzzy sets for each of types. We compared the crisp transitions obtained from our previous model with the proposed fuzzy transitions on both computer generated clusters and human annotated topic sets. Both crisp and fuzzy transition charts showed consistency on how \textit{strong} transitions progress overtime.

Our results also suggest that fuzzy transitions demonstrate evolution of topics at varying granularity: \textit{strong} set transitions can trace topics that have a strong presence in the clusters, while \textit{weaker} set transition pay attention to smaller details. Depending on the level of details that one wishes to concentrate, it may or may not be necessary to use fuzzy transitions.  However, if one wishes to trace topics when they emerge, even weakly, fuzzy transitions seem to provide value.  

\begin{acknowledgement}
“This is a preprint of the accepted manuscript: Xiaonan Jing, Yi Zhang, Qingyuan Hu, Julia Taylor Rayz, Modeling Fuzzy Cluster Transitions for Topic Tracing, published in NAFIPS'2021, edited by Julia Rayz, Victor Raskin, Scott Dick, and Vladik Kreinovich, 2021, Explainable AI and Other Applications of Fuzzy Techniques reproduced with permission of Springer Nature Switzerland AG. The final authenticated version is available online at: TBD”.
\end{acknowledgement}

\bibliography{references}

\begin{thebibliography}{10}
\providecommand{\url}[1]{{#1}}
\providecommand{\urlprefix}{URL }
\expandafter\ifx\csname urlstyle\endcsname\relax
  \providecommand{\doi}[1]{DOI~\discretionary{}{}{}#1}\else
  \providecommand{\doi}{DOI~\discretionary{}{}{}\begingroup
  \urlstyle{rm}\Url}\fi

\bibitem{guille2015event}
Guille, A., Favre, C.: Event detection, tracking, and visualization in twitter:
  a mention-anomaly-based approach.
\newblock Social Network Analysis and Mining \textbf{5}(1), 18 (2015)

\bibitem{meladianos2015degeneracy}
Meladianos, P., Nikolentzos, G., Rousseau, F., Stavrakas, Y., Vazirgiannis, M.:
  Degeneracy-based real-time sub-event detection in twitter stream.
\newblock Proceedings of the International AAAI Conference on Web and Social
  Media \textbf{9}(1) (2015)

\bibitem{fedoryszak2019real}
Fedoryszak, M., Frederick, B., Rajaram, V., Zhong, C.: Real-time event
  detection on social data streams.
\newblock In: Proceedings of the 25th ACM SIGKDD International Conference on
  Knowledge Discovery \& Data Mining, pp. 2774--2782 (2019)

\bibitem{zadeh1965}
Zadeh, L.: Fuzzy sets.
\newblock Information and Control \textbf{8}(3), 338--353 (1965).
\newblock \doi{https://doi.org/10.1016/S0019-9958(65)90241-X}

\bibitem{jing2021tracing}
Jing, X., Hu, Q., Zhang, Y., Rayz, J.R.: Tracing topic transitions with
  temporal graph clusters.
\newblock In proceedings of the Thirty-Fourth International FLAIRS Conference
  (FLAIRS-34)  (2021)

\bibitem{van2008graph}
Van~Dongen, S.: Graph clustering via a discrete uncoupling process.
\newblock SIAM Journal on Matrix Analysis and Applications \textbf{30}(1),
  121--141 (2008)

\bibitem{alnajran_cluster_2017}
Alnajran, N., Crockett, K., McLean, D., Latham, A.: Cluster {Analysis} of
  {Twitter} {Data}: {A} {Review} of {Algorithms}:.
\newblock In: Proceedings of the 9th {International} {Conference} on {Agents}
  and {Artificial} {Intelligence}, pp. 239--249. SCITEPRESS - Science and
  Technology Publications, Porto, Portugal (2017).
\newblock \doi{10.5220/0006202802390249}

\bibitem{zadeh2015fuzzy}
Zadeh, L., Abbasov, A., Shahbazova, S.N.: Analysis of twitter hashtags: Fuzzy
  clustering approach.
\newblock 2015 Annual Conference of the North American Fuzzy Information
  Processing Society (NAFIPS) held jointly with 2015 5th World Conference on
  Soft Computing (WConSC) pp. 1--6 (2015).
\newblock \doi{10.1109/NAFIPS-WConSC.2015.7284196}

\bibitem{spiliopoulou_monic_2006}
Spiliopoulou, M., Ntoutsi, I., Theodoridis, Y., Schult, R.: {MONIC}: {Modeling}
  and {Monitoring} {Cluster} {Transitions}.
\newblock In: Proceedings of the 12th {ACM} {SIGKDD} {International}
  {Conference} on {Knowledge} {Discovery} and {Data} {Mining}, {KDD}'06, pp.
  706--711. Association for Computing Machinery, New York, NY, USA (2006).
\newblock \doi{10.1145/1150402.1150491}.
\newblock Event-place: Philadelphia, PA, USA

\bibitem{oliveira_framework_2012}
Oliveira, M., Gama, J.: A framework to monitor clusters evolution applied to
  economy and finance problems.
\newblock Intelligent Data Analysis \textbf{16}(1), 93--111 (2012).
\newblock \doi{10.3233/IDA-2011-0512}

\bibitem{gong_clustering_2017}
Gong, S., Zhang, Y., Yu, G.: Clustering {Stream} {Data} by {Exploring} the
  {Evolution} of {Density} {Mountain}.
\newblock Proceedings of the VLDB Endowment \textbf{11}(4), 393--405 (2017).
\newblock \doi{10.1145/3164135.3164136}.
\newblock Publisher: Very Large Data Bases Endowment

\end{thebibliography}
\bibliographystyle{spmpsci}

\end{document}